\documentclass{article}
\pdfoutput=1
\usepackage[preprint]{spconf}
\usepackage{amsmath,graphicx}
\usepackage{booktabs}

\title{Non-iterative optimization of pseudo-labeling thresholds for training
  object detection models from multiple datasets}
\name{Yuki Tanaka, Shuhei M.~Yoshida, Makoto Terao\thanks{\copyright\ 2022
    IEEE. Personal use of this material is permitted. Permission from IEEE must
    be
    obtained for all other uses, in any current or future media, including
    reprinting/republishing this material for advertising or promotional
    purposes,
    creating new collective works, for resale or redistribution to servers or
    lists, or reuse of any copyrighted component of this work in other works.}}
\address{Visual Intelligence Research Laboratories\\
  NEC Corporation\\
  Kawasaki, Kanagawa, Japan}
\copyrightnotice{\copyright\ IEEE 2022}
\begin{document}
\maketitle
\begin{abstract}
  We propose a non-iterative method to optimize pseudo-labeling thresholds for
  learning object detection from a collection of low-cost datasets, each of
  which
  is annotated for only a subset of all the object classes.
  A popular approach to this problem is first to train teacher models and then
  to
  use their confident predictions as pseudo ground-truth labels when training a
  student model.
  To obtain the best result, however, thresholds for prediction confidence must
  be adjusted.
  This process typically involves iterative search and repeated training of
  student models and is time-consuming.
  Therefore, we develop a method to optimize the thresholds without iterative
  optimization by maximizing the $F_\beta$-score on a validation dataset, which
  measures the quality of pseudo labels and can be measured without training a
  student model.
  We experimentally demonstrate that our proposed method achieves an mAP
  comparable to that of grid search on the COCO and VOC datasets.
\end{abstract}
\section{Introduction}
\label{sec:intro}

Object detection~\cite{faster-rcnn, SSD, yolo, m2det} has achieved significant
progress in deep learning with a tremendous number of images and annotations,
but it becomes quite expensive to collect them. This creates a significant
barrier when it comes to moving from the research stage to practical
application. Recently, research on how to train a model with low-cost datasets
has become more active.

\begin{figure}[t]
  \begin{minipage}[b]{0.49\linewidth}
    \centering

    \centerline{\includegraphics[width=0.9\linewidth]{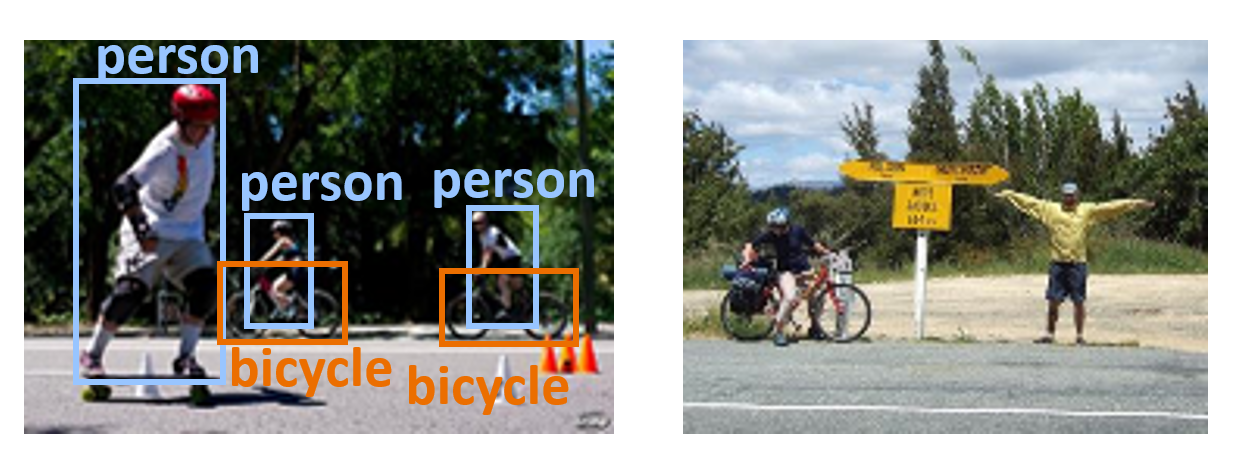}}
    \centerline{(a) Semi-supervised learning}\medskip
  \end{minipage}
  \hfill
  \begin{minipage}[b]{0.49\linewidth}
    \centering

    \centerline{\includegraphics[width=0.9\linewidth]{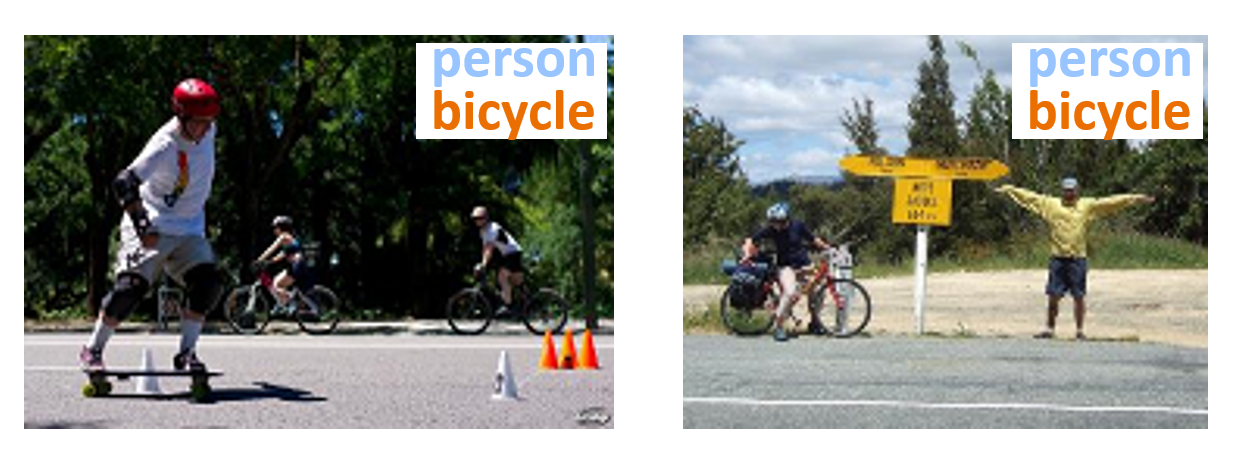}}
    \centerline{(b) Weakly supervised learning}\medskip
  \end{minipage}

  \begin{minipage}[b]{1\linewidth}
    \centering

    \centerline{\includegraphics[width=0.9\linewidth]{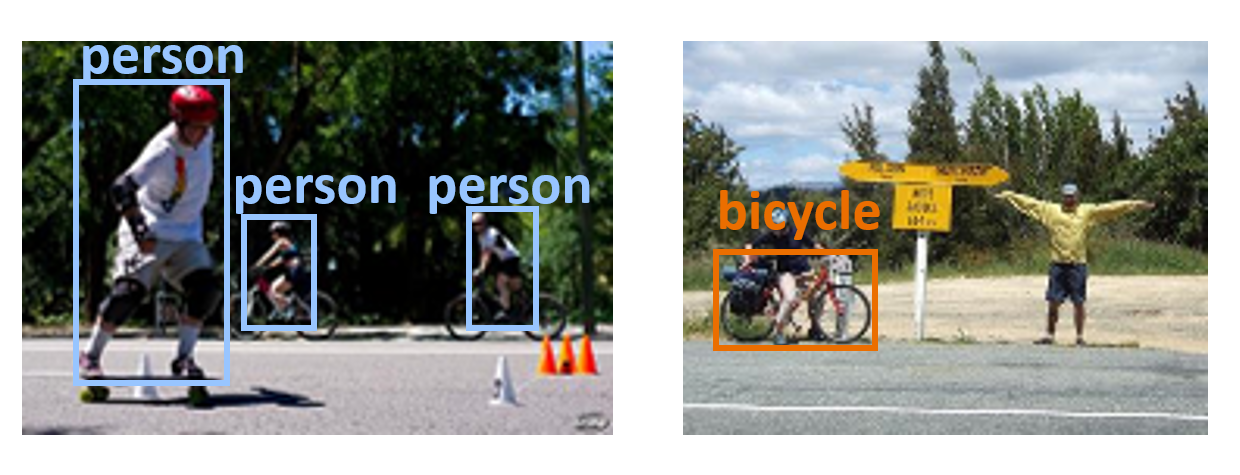}}
    \centerline{(c) Learning from multiple datasets with different class
      sets}\medskip
  \end{minipage}
  \caption{Examples of learning paradigms for object detection using low-cost
    datasets. In these examples, the goal is to train a model that detects
    people
    and bicycles in images by using training datasets with annotations as
    illustrated above. The two images are contained in COCO~\cite{COCO} and
    VOC~\cite{VOC} datasets, respectively.}
  \label{fig:dataset overview}
\end{figure}

There are several paradigms to learn from low-cost datasets.
Examples include semi-supervised learning~\cite{SSL_CR_object_detection, STAC,
  softteacher_SSL,interpolation_SSL} and weakly supervised
learning~\cite{WSOD_2016, WSOD_PCL}.
In semi-supervised learning, models are trained from a limited amount of
labeled data and a lot of unlabeled data (Fig.~\ref{fig:dataset overview}(a)),
while in weakly supervised learning, models are trained from only image-level
annotations and no bounding boxes (Fig.~\ref{fig:dataset overview}(b)).
By contrast, we aim at training a single object detection model for all classes
from multiple datasets that have different class sets without additional
annotations~\cite{Zhao_UniDet_ECCV20,zhao2020towards, yao2020cross}.
This setting~(Fig.~\ref{fig:dataset overview}(c)) is important for practical
applications, because we can add object categories simply by combining datasets
that are made for different purposes.

\begin{figure*}
  \centering
  \includegraphics[width=0.9\linewidth]{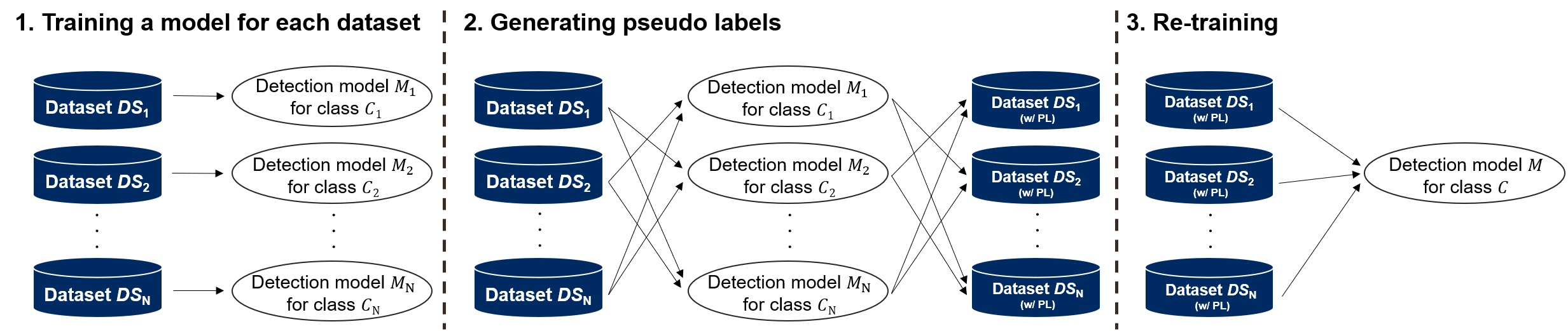}
  \caption{Schematic picture of the pseudo-labeling method for training an
    object
    detection model from multiple datasets with different class sets.
    First, a dataset $DS_i$ is used to train a model $M_i$.
    Then the model $M_i$ generates pseudo labels of the object classes in $C_i$
    for
    a dataset $DS_j$ ($i\neq j$).
    Finally, the datasets with pseudo labels are used to train the final
    detection
    model $M$.
  }
  \label{fig:pipeline of pseudo label method}
\end{figure*}
Typically, pseudo labeling is used to train an object detection model in the
current problem setting (Fig.~\ref{fig:pipeline of pseudo label method}).
Specifically, we first train one teacher model from each dataset and then use
them to predict locations of unlabeled objects.
A prediction is used as a pseudo label if its confidence score is higher than a
predetermined threshold.
Finally, we train a single student model for all classes by using both the
ground-truth labels and the pseudo labels.

To achieve the best performance with pseudo labeling, it is imperative to
decide this threshold properly, but the optimization process is time-consuming.
This optimization usually involves iterative search; we generate pseudo labels,
train and evaluate a student model, and then repeat these steps multiple times.
If we use a common value for all classes, this might require, e.g., 10
repetitions.
Moreover, if we wish to find an optimal threshold for each of $K$ classes,
the naive grid search algorithm requires $10^{K}$ iterations, which becomes
infeasible if $K$ is more than a few.

In this work, we propose a non-iterative method for optimizing thresholds for
pseudo labeling.
We determine the thresholds so that the $F_\beta$-score of pseudo labels or
that of both ground-truth and pseudo labels is maximized; see
Sections~\ref{ssec:Fscore_PL} and \ref{ssec:PL_DS} for the definitions of these
quantities.
Importantly, these metrics can be measured without training a student model,
unlike the detection performance of a student model on a validation dataset,
which is usually used as a criterion to find the optimal thresholds.
With our proposed method, the training of a student model is required only
once, when we obtain the final detection model.
We experimentally demonstrate that our proposed method can achieve a comparable
mAP to that of the conventional grid search method, which involves repeated
training of student models.

\section{Method}
\label{sec:method}
We provide a brief overview of pseudo labeling in Section~\ref{ssec:PL} and
then explain our proposed method for optimizing pseudo-labeling thresholds in
the following subsections.

\subsection{Pseudo labeling}
\label{ssec:PL}

In Fig.~\ref{fig:pipeline of pseudo label method}, we illustrate the
pseudo-labeling method for learning object detection from multiple datasets
with different sets of object classes.
Assume that the entire dataset $DS$ consists of $N$ datasets $DS_{1},
  DS_{2},\dots, DS_{N}$ that have different class sets $C_{1}, C_{2}, \dots,
  C_{N}$, respectively. The class sets of dataset $DS$ can be written as $C =
  C_{1} \cup C_{2} \cup \dots \cup C_{N}$, and a complementary label set of
dataset $DS_{i}$ can be written as $\overline{C_{i}} = (C \setminus C_{i})$.
First, we train an object detection model $M_{i}$ for a label set $C_{i}$ with
a dataset $DS_{i}$ by supervised learning.
We train models with $N$ datasets individually, so $N$ object detection models
are generated.
Second, we generate pseudo labels.
To this end, models that cover $\overline{C_i}$ are used to get predictions of
object locations in the dataset $DS_i$.
If the raw predictions were used as pseudo labels, they would be too noisy, so
we only adopt predictions with the confidence scores higher than a certain
threshold.
Finally, we train an object detection model $M$ with the original ground-truth
labels and the generated pseudo labels.

There is a variant of the pseudo-labeling method where two thresholds, $\tau_h$
and $\tau_l$, are used to further reduce the noise of pseudo
labels~\cite{Zhao_UniDet_ECCV20}.
If a prediction has a confidence score higher than $\tau_h$, then it is used as
a pseudo label, while if the confidence scores of all the classes in a region
are below $\tau_l$, such an area is treated as pseudo background.
Regions that do not satisfy either are ignored in the training of a student
model.

\subsection{Maximizing $F_\beta$-score of pseudo labels}
\label{ssec:Fscore_PL}

The goal of optimizing the thresholds is to find a set of pseudo labels that
brings us the best student model.
However, it is time-consuming to measure the performance of a student model for
each pseudo-labeled dataset generated with a different threshold.
Therefore, we propose to use $F_\beta$-score of pseudo labels as a surrogate
for the student performance.
It is defined as the weighted harmonic mean of precision and recall:
\begin{equation}
  F_\beta=\frac{(1+\beta^2) \cdot precision \cdot recall}{\beta^2precision +
    recall}.
\end{equation}
If the $\beta$ value is less than 1, $F_\beta$ is a precision-weighted metric,
while if the $\beta$ value is more than 1, $F_\beta$ is a recall-weighted
metric.
Importantly, because this essentially measures performance of the teacher
models, we do not need to train a student model to calculate the
$F_\beta$-score.

Ideally we would evaluate the $F_\beta$-score of the pseudo labels themselves,
but this is impossible because there are no ground-truth labels in the dataset
that we want to generate pseudo labels for.
Therefore, we measure the $F_\beta$-score on a pseudo-label validation dataset.
We can prepare this dataset either by splitting from a training dataset or
separately under the condition that it has annotations of evaluation
categories.

To find the optimal thresholds, we use the teacher models $M_i$ to generate
predictions of object locations in the validation dataset and calculate the
$F_\beta$-score.
For the one-threshold variant of pseudo labeling, we adopt the threshold that
brings the maximum $F_{1}$-score as $\tau$.
For the two-threshold variant, on the other hand, we take as $\tau_{h}$ the
threshold that maximizes the $F_{0.5}$-score, a typical precision-weighted
metric, and as $\tau_{l}$ the threshold that maximizes the $F_{2}$-score, a
typical recall-weighted metric.

\subsection{Maximizing $F_\beta$-score of all labels}
\label{ssec:PL_DS}

\begin{figure}[t]
  \centering
  \includegraphics[width=0.9\linewidth]{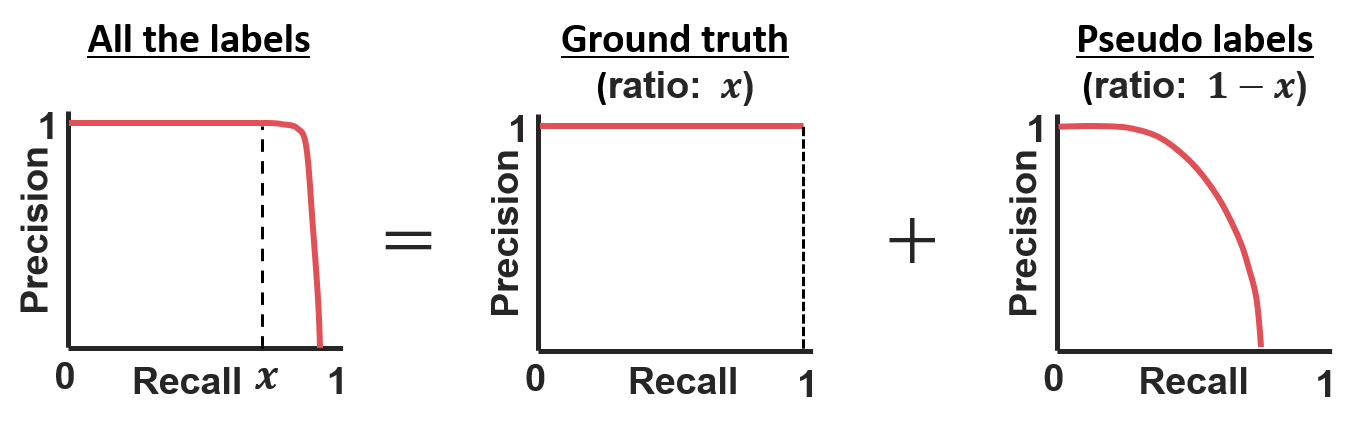}
  \caption{Schematic image of a precision-recall curve of all labels including
    both ground-truth and pseudo labels. Thresholds are determined using not
    only
    automatically generated pseudo labels but also human-annotated ground-truth
    labels.}
  \label{fig:overview of Fmax_DS}
\end{figure}
The method discussed above determines thresholds by maximizing the
$F_\beta$-score of pseudo labels.
However, because training of a student model uses both the human-annotated
ground-truth labels and the generated pseudo labels, it is preferable to take
both of them into account to determine the thresholds.

Figure~\ref{fig:overview of Fmax_DS} illustrates the basic idea. For the object
class $j$, let $x_{j}$ be the ratio
$
  \frac{g_{j}}{t_{j}}
$
of the number $g_j$ of labeled objects over the total number $t_j$ of labeled
and unlabeled ones.
We assume that the ground-truth labels are definitely correct, so we set
precision = 1 and confidence score = 1.
Then we can calculate the precision and recall of all the labels as
\begin{align}
  p_{DS_{j}}(\tau) & = \left \{
  \begin{array}{ll}
    1                                     & (\tau = 1), \\
    \frac{x_{j}t_{j} + (1 - x_{j})t_{j}r_{j}(\tau)}{x_{j}t_{j} + \frac{(1 -
    x_{j})t_{j}r_{j}(\tau)}{p_{j}(\tau)}} & (\tau < 1),
  \end{array}
  \right.                       \\
  r_{DS_{j}}(\tau) & = \left \{
  \begin{array}{ll}
    \frac{g_{j}}{t_{j}}            & (\tau = 1), \\
    x_{j} + (1 - x_{j})r_{j}(\tau) & (\tau < 1).
  \end{array}
  \right.
\end{align}
Here, $p_j(\tau)$ and $r_j(\tau)$ are precision and recall of pseudo labels
only.
By using these expressions, we can calculate the $F_\beta$-score of all labels
and select thresholds $\tau$, or the pair $\tau_{h}$ and $\tau_{l}$, in the
same way as explained in Section~\ref{ssec:Fscore_PL}.

Under the current problem setting where no knowledge about unlabeled instances
is available, $t_{j}$ is also unknown, because this includes the number of
unlabeled objects.
In such cases, a crude but simple way to estimate $x_{j}$ is to use the ratios
of the number of images in the datasets with annotations of the object class
$j$ and the number of images in the entire dataset.

\section{Experiments}

In this section, we experimentally demonstrate the effectiveness of our
proposed method.
Here, we compare four methods including a variant of our proposed method.
``w/o PL'' only uses the ground-truth annotation without pseudo labels; this
sets the lower-bound performance for the other methods.
``Grid search'' finds the optimal threshold from a predefined pool of candidate
values.
With this method, we use the same threshold value for all the classes because
otherwise it is infeasible to perform grid search due to the enormous search
space.
``$\rm{Fmax_{PL}}$'' and ``$\rm{Fmax_{DS}}$'' are our proposed methods.
The former uses the $F_\beta$-scores of pseudo labels to determine the optimal
thresholds (Section~\ref{ssec:Fscore_PL}), while the latter uses those of all
the labels including both ground-truth and pseudo labels
(Section~\ref{ssec:PL_DS}).

\subsection{Datasets}

We experiment with two semi-synthetic datasets, COCO-splitting and COCO-VOC,
which emulate the problem setting as described in Section~\ref{sec:intro}.

\textbf{COCO-splitting}
We split Microsoft COCO Detection 2017 (COCO)~\cite{COCO}, a general object
detection dataset consisting of 80 object classes, into $N$ subsets ($N = 2, 5,
  10$) to make multiple datasets with different class sets, as follows:
split 110,000 images from COCO training data into $N$ subsets to make $N$
datasets ($DS_{i}$);
assign category ids $i$ (mod $N$) to $C_i$; and
remove images in $DS_{i}$ that have no objects belonging to a category in
$C_i$.
In this setting, the $N$ splits are expected to have similar characteristics
with each other.
The remaining 5,000 images of the COCO training split are used as a
pseudo-label validation dataset.

\textbf{COCO-VOC}
This dataset combines COCO~\cite{COCO} and Pascal VOC~\cite{VOC} (COCO-VOC
dataset), mimicking the situation where there is a domain gap between
constituent datasets.
The VOC dataset consists of 20 object categories, which is a subset of COCO's
80 classes.
We remove annotations of the overlapping classes from COCO, so that the two
datasets have mutually exclusive sets of object classes.
Similarly to COCO-splitting, we use 5,000 images from COCO's training split and
500 images from VOC's trainval split for validation.
For testing, we use the same dataset as used in Ref.~\cite{Zhao_UniDet_ECCV20},
which is taken from VOC's test set and COCO's validation set.
The VOC portion of this dataset annotates not only 20 object categories but
also an additional 60 categories that are only annotated in the COCO training
dataset.
This allows us to measure mAP over 80 classes on both of the datasets.

\subsection{Implementation details}

We use M2Det~\cite{m2det} as an object detection model.
M2Det is a one-stage object detection model that applies a multi-level feature
pyramid network (MLFPN) to a feature map generated from a backbone to perform
detection by using multi-scale feature maps.
The input image size is $320 \times 320$ pixels and a VGG16 network pretrained
on ImageNet is used as the backbone.
For each experiment, the model is trained for 150 epochs by using NesterovAG
with an initial learning rate of 0.01 and a momentum of 0.9.

In the experiment with COCO-splitting, we use the simpler version of the
pseudo-labeling method with one threshold to reduce the computational cost.
We also found from preliminary experiments that the gain from the second
threshold was tiny for this dataset.
On the other hand, we adopt the two-threshold variant with COCO-VOC, because
preliminary experiments suggested that this setting benefit significantly from
the second, lower threshold.

For the grid search algorithm, we need to predetermine the sets of candidate
threshold values.
When tuning the single threshold $\tau$, we take $[0.2, 0.3, 0.4, 0.5, 0.6,
      0.7, 0.8, 0.9, 1]$ as the pool of the candidates.
Note that taking the value $\tau=1$ is equivalent to ``w/o PL'', which simply
combines datasets without any additional pseudo labels.
On the other hand, the thresholds $\tau_{h}$ and $\tau_{l}$ are selected from
$[0.2, 0.4, 0.6, 0.8, 1.0]$, with the constraint $\tau_h \geq \tau_l$.
This results in 15 pairs of the candidate threshold values for this algorithm.

In the $\rm{Fmax_{DS}}$ method, we need to estimate the ratio of the number of
labeled objects to that of all objects for each object class.
In the COCO splitting experiment, for simplicity, we use the true number of
$x_{j}$, which is known from COCO's original annotation.
On the other hands, with COCO-VOC, we follow the simple procedure explained in
Section~\ref{ssec:PL_DS}.

\subsection{Results}

\begin{figure}
  \centering
  \includegraphics[width=0.9\linewidth]{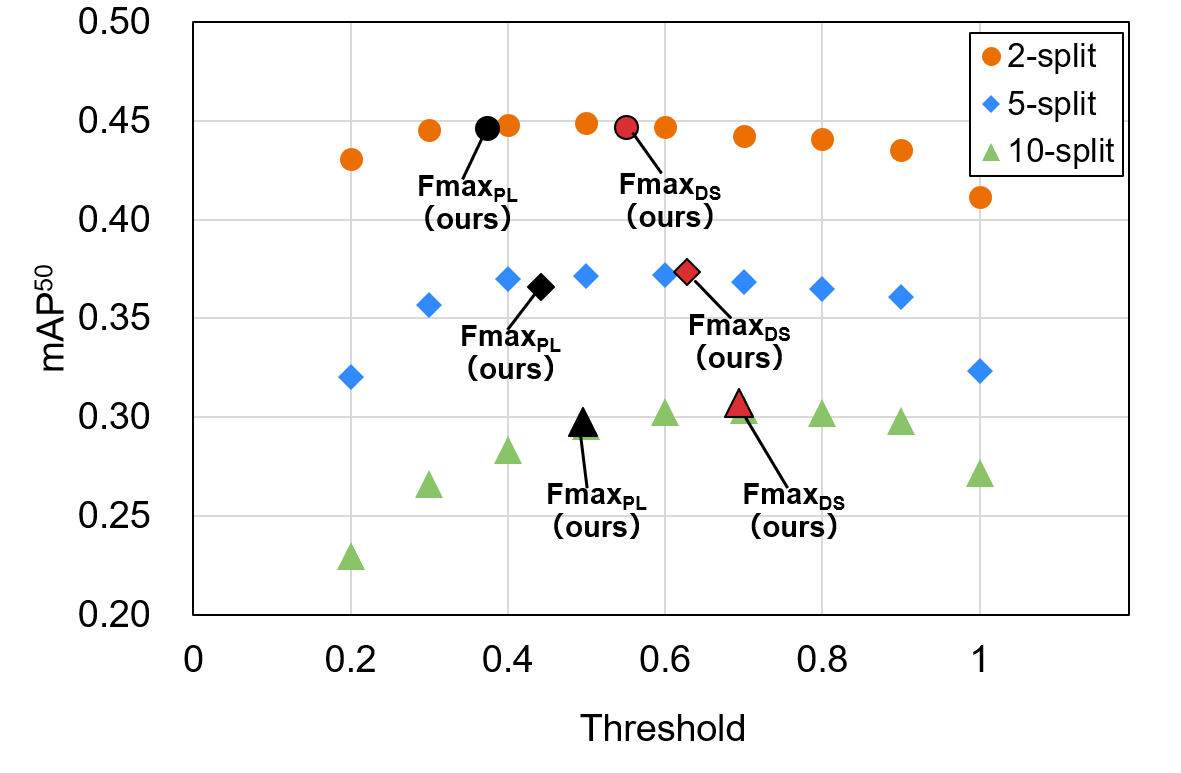}
  \caption{Results of COCO-splitting dataset. We compare our proposed methods
    ($\rm{Fmax_{PL}}$ and $\rm{Fmax_{DS}}$) with grid search.
    Note that grid search with the threshold equal to one is equivalent to
    ``w/o
    PL''.
    The thresholds for the proposed methods are the average over the thresholds
    for
    the 80 classes.}
  \label{fig:result of COCO}
\end{figure}
\begin{table}
  \centering
  \caption{Results of COCO-splitting dataset. We show mAP$^{50}$ for ``w/o
    PL'', grid search, and our proposed methods.}
  {\small
    \begin{tabular}{l|cccc}
      \toprule
                       & 2 splits
                       & 5 splits
                       & 10 splits         \\
      \midrule%
      w/o PL           &
      0.411            & 0.323     & 0.272 \\
      Grid search      &
      0.449            & 0.372     & 0.304 \\
      $\rm{Fmax_{PL}}$ &
      0.446            & 0.366     & 0.298 \\
      $\rm{Fmax_{DS}}$ &
      0.447            & 0.373     & 0.307 \\
      \bottomrule
    \end{tabular}
  }
  \label{tab:coco-splitting}
\end{table}

Figure~\ref{fig:result of COCO} shows the experimental results for the
COCO-splitting datasets.
The plot indicates that one of our proposed method, $\rm{Fmax_{DS}}$, performs
better than or competitively to the grid search method, without any iterative
search for the optimal $\tau$.
$\rm{Fmax_{PL}}$ prefers smaller thresholds, which means that it uses noisier
pseudo labels to train a student model.
This slightly degrades the detection accuracy, as compared with grid search and
$\rm{Fmax_{DS}}$, but it is still competitive~(Table~\ref{tab:coco-splitting}).

Interestingly, Fig.~\ref{fig:result of COCO} implies that the optimal value of
the threshold depends on a dataset.
In other words, there is no universal value of the best threshold that leads to
the optimal performance for any dataset.
This observation corroborates the underlying premise of our research that we
need to optimize the threshold to achieve the best results.
Figure~\ref{fig:result of COCO} also indicates that the value found by
$\rm{Fmax_{DS}}$ is similar to the optimal value found by grid search.

\begin{table}
  \centering
  \caption{Results of COCO-VOC dataset. We show mAP$^{50}$ for ``w/o PL'',
    grid search, and our proposed methods.
    ``C+V'' denotes the evaluation on the combined test dataset of COCO and
    VOC, while ``C'' and ``V'' are the COCO and VOC portions of the test
    dataset,
    respectively.
    $\tau_h$ and $\tau_l$ are the higher and lower thresholds chosen by each
    method.
  }
  {\small
    \begin{tabular}{l|ccccc}
      \toprule
                       & {C+V}
                       & {C}
                       & {V}
                       & {$(\tau_{h},\tau_{l})$}                          \\
      \midrule%
      w/o PL           &
      0.425            & 0.422                   & 0.343 & ---            \\
      Grid search      &
      0.480            & 0.489                   & 0.414 & $(0.8, 0.2)$   \\
      $\rm{Fmax_{PL}}$ &
      0.478            & 0.482                   & 0.431 & $(0.61, 0.24)$ \\
      $\rm{Fmax_{DS}}$ &
      0.481            & 0.483                   & 0.422 & $(0.82,0.29)$  \\
      \bottomrule
    \end{tabular}
  }
  \label{tab:voccoco}
\end{table}

Table~\ref{tab:voccoco} lists the experimental results of the COCO-VOC dataset.
Without repeated training of student models,
$\rm{Fmax_{DS}}$ achieves the best result with respect to mAP over the whole
test set (``C+V'').
This result suggests that our proposed method is robust to the domain gap
between the constituent datasets.

\section{Conclusion}
\label{sec:conc}

In this paper, we proposed a non-iterative method to optimize pseudo-labeling
thresholds for training a single object detection model from multiple datasets.
To avoid training a student model multiple times, we used $F_\beta$-score to
measure the quality of pseudo labels and to find the optimal threshold.
Experimental results showed that our method achieved an mAP competitive with
grid search, but with significantly lower computational costs.
This work should prove helpful for the implementation of deep learning to
practical applications at low cost.

\bibliographystyle{IEEEbib}
\bibliography{refs}

\end{document}